\pdfoutput=1
\PassOptionsToPackage{usenames,dvipsnames}{xcolor}

\documentclass[11pt]{article}

\usepackage[]{acl}
\usepackage[linesnumbered, ruled, vlined]{algorithm2e}
\usepackage{times}
\usepackage{latexsym}
\usepackage{soul}
\usepackage[T1]{fontenc}
\usepackage{amsmath,amsfonts}
\usepackage{booktabs}
\usepackage{footnote}
\usepackage[utf8]{inputenc}
\usepackage{graphicx}
\usepackage{microtype}
\title{PTP: Boosting Stability and Performance of Prompt Tuning with Perturbation-Based Regularizer}

\newcommand{\hld}[1]{\textcolor{red}{ \scriptsize #1}}
\newcommand{\down}[1]{\textcolor{blue}{\scriptsize #1}}
\author{
Lichang Chen \\ 
University of Maryland \\ 
bobchen@umd.edu \\ \And
Heng Huang\\ 
University of Maryland \\ 
heng@umd.edu \\ \And
  Minhao Cheng \\
  HKUST \\
  minhaocheng@ust.hk
  }

\begin{document}
\maketitle
\begin{abstract}
Recent studies show that prompt tuning can better leverage the power of large language models than fine-tuning on downstream natural language understanding tasks. However, the existing prompt tuning methods have training instability issues, as the variance of scores under different random seeds is quite large. 
To address this critical problem, we first investigate and find that the loss landscape of vanilla prompt tuning is precipitous when it is visualized, where a slight change of input data can cause a big fluctuation in the loss landscape. This is an essential factor that leads to the instability of prompt tuning. %\cmh{a little more explanation} 
Based on this observation, we introduce perturbation-based regularizers, which can smooth the loss landscape, into prompt tuning. We propose a new algorithm, called Prompt Tuning with Perturbation-based regularizer~(PTP), which can not only alleviate training instability dramatically but also boost the performance of prompt tuning. We design two kinds of perturbation-based regularizers, including random-noise-based and adversarial-based. In particular, our proposed perturbations are flexible on both text space and embedding space. Extensive experiments show the effectiveness of our proposed methods in stabilizing the training. Our new algorithms improve the state-of-the-art prompt tuning methods by 1.94\% and 2.34\% on SuperGLUE and FewGLUE benchmarks, respectively.
% To the best of our knowledge, we are the first to adapt adversarial training to the prompt tuning. 
% illustrate the advantage of prompt tuning 
\end{abstract}

\section{Introduction}
% introduce adversarial training in the 1st paragraph. LMs and prompt tuning -> 2nd paragraph
% \par Releasing the burden of training models from scratch while keeping the outstanding performance on downstream tasks, pretrained Language Models~(LMs) brought NLP to a new era.~\cite{raffel2020exploring, he2020deberta}. 
% in order to make the with the parameters of LMs increasing to 10B or even 100B, finetuning with such a large pretrained LMs may cause GPU out of memory problem~\cite{megaron-LM, gpt3, ptuning2021}. %performance --> 
% Fine-tuning Large Models~(LMs) with classification head becomes a trend since BERT~\cite{BERT} appears, where the backbone is not frozen and all the parameters are tunable. However, it is memory-consuming to store all the parameters and their gradient and, at the same time, inconvenient to fine-tune the pretrained model on every single downstream task separately.

% introduce P-tuning as prompt-based finetuning method and introduce PT2 as prompt tuning methods. Why few-shot and fully-supervised setting in your experiment.
\par Releasing the burden of training models from scratch while keeping the outstanding performance on downstream tasks, pretrained Language Models~(LMs) brought NLP to a new era~\cite{raffel2020exploring, he2020deberta, megaron-LM}. Since BERT~\cite{BERT}, fine-tuning all the parameters of pretrained LMs becomes a common practice. However, it is memory-consuming to store a copy of the entire LM for each downstream task due to the number of parameters in LM can be 10B or even 100B~\cite{megaron-LM, gpt3}.
\begin{figure}[t]
    \centering
    \includegraphics[width=0.49\textwidth]{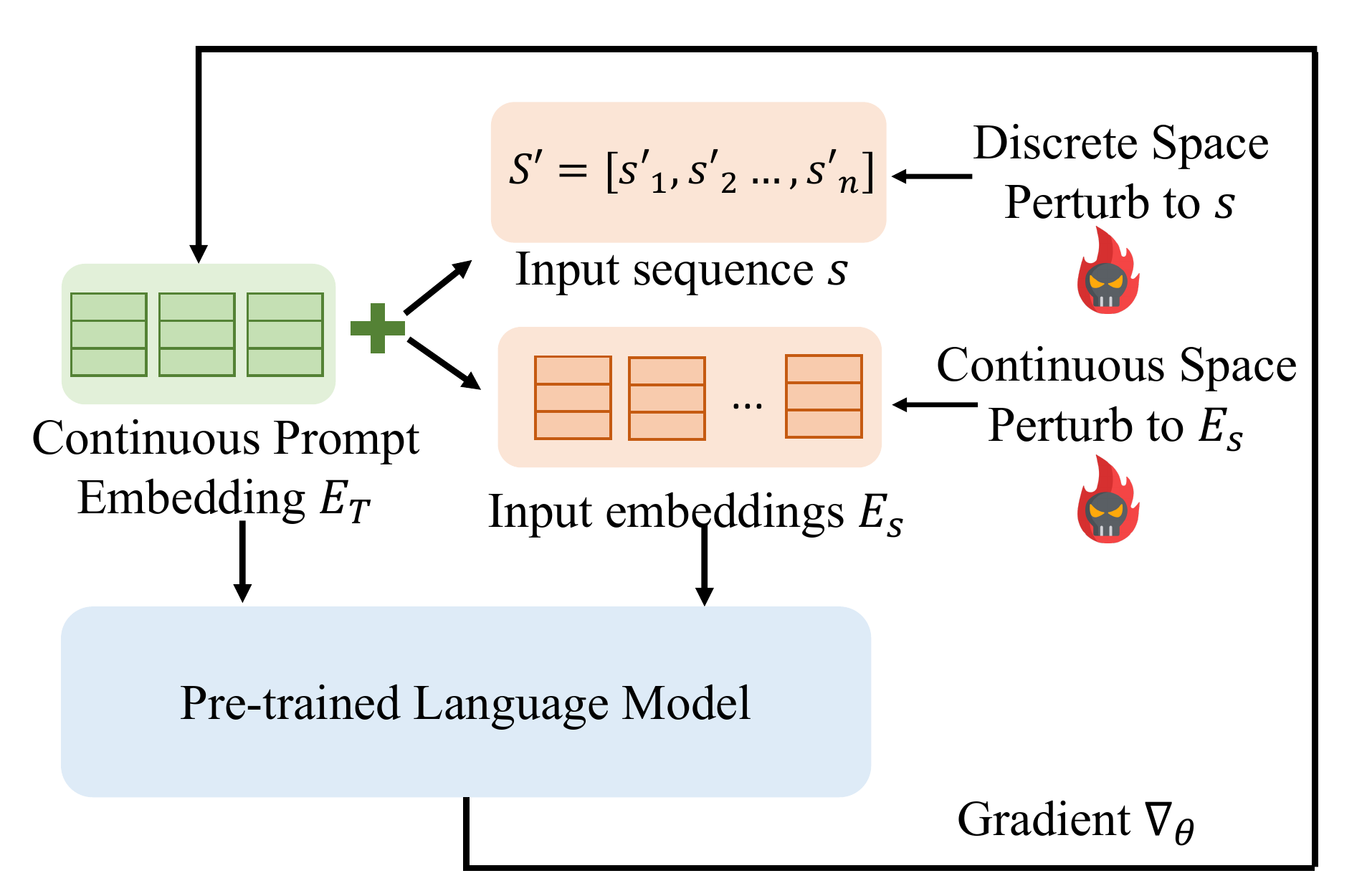}
    \caption{A simple pipeline of our \textbf{P}ropmt \textbf{T}uning with \textbf{P}erturbation-based regularizor~(PTP) algorithm. 
    %Through the perturbations on continuous space or discrete space, we obtain the perturbed data and we train it together with clean data in the prompt tuning paradigm.
    }
    \label{fig:teaser}
    % arrow to show adversarial training
\end{figure}

\par Recently, inspired by the success of GPT-3~\cite{gpt3} on few-shot and zero-shot learning with manually created prompts, there has been a surging interest in prompting that freezes pre-trained LM and wraps up the input sequence with natural language templates. However, natural language prompts are handcrafted by experts and the performance is not comparable with fine-tuning methods. To tackle it, \citet{lester2021power,li2021prefix} proposed prompt tuning, which prepends the input sequence with continuous embeddings and only tunes these embeddings during training. \citet{ptuning2021,ptuningv2} verified the effectiveness of prompt tuning on natural language understanding~(NLU) tasks under both few-shot learning and supervised learning settings, which is comparable to the fine-tuning methods but with much fewer~($1000\times$ less) task-specific tunable parameters. However, under different random seeds, we observe that the current prompt tuning methods suffer from a high variance of scores, which indicates they suffer from training instability issues.

% Inspired by the success of GPT-3~\cite{gpt3}, there has been a rapidly growing interest in its new pretraining and prompting paradigm, which gradually substitute the tradition pretraining then fine-tuning paradigm~\cite{schick2020exploiting, gao2020making, fewglue}.  

% GPT-3~\cite{gpt3} proposes a new way, by wrapping up the input sequence with a template, to leverage LMs better and achieve outstanding results in downstream tasks with just a few of examples. These magic templates are composed of natural~(English) words and known as prompts.~\cite{gpt3, schick2020exploiting, gao2020making, fewglue}. However, searching for the appropriate discrete prompt is time-consuming and labor-intensive~\cite{DavisonFR19}. As the goal of prompts is to assist LM in performing a task effectively, the interpretablity is not necessary. Therefore, continuous prompt tuning methods~\cite{li2021prefix, HambardzumyanKM20} adopt trainable embeddings to replace the natural words of prompts. Then the embeddings are tuned by gradient directly. With much fewer trainable parameters~(10x or even 100x fewer), it can achieve a comparable or even better performance than the classical approach.~\cite{ptuning2021, ptuningv2, lester2021power}. % how to do continuous prompt tuning

\begin{figure*}[htp]
    \centering
    \includegraphics[width=0.95\textwidth]{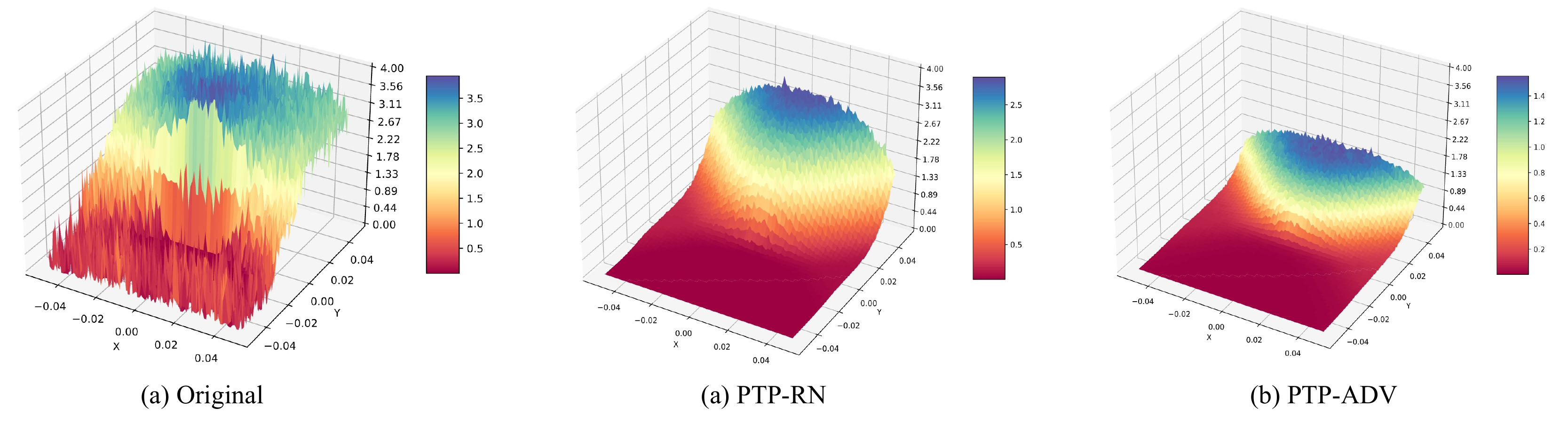}
    \caption{The loss landscapes on different continuous prompt tuning procedures. The X-axis and Y-axis denote the magnitude of perturbations~(gradient direction) and another random orthogonal direction. Z-axis represents the cross-entropy loss of the different training methods. }
    %\cmh{which loss? what loss?} }
    \label{fig:loss landscape}
    % arrow to show adversarial training
\end{figure*}

% do not mention xiangning's paper
\par To investigate the factor that causes the instability of prompt tuning, we visualize the loss landscape of the vanilla prompt tuning and observe that there exist many sharp crests, as shown in Figure~\ref{fig:loss landscape}(a), which  harms the training stability. Motivated by the recent study~\cite{chen20perturb} which shows that perturbation-based regularizers are powerful tools to smooth the loss landscape and stabilize the training of machine learning systems, we introduce them into prompt tuning to address the lack of stability and generalization issues. To be specific, we propose Prompt Tuning with Perturbation-based regularizer~(PTP) algorithm to make the training stable and boost the performance of prompt tuning. 

\par Specifically, we consider two kinds of perturbations in PTP, Random-Noise-based~(PTP-RN) perturbation and ADVersarial-based~(PTP-ADV) perturbation. PTP-RN is motivated by randomized smoothing~\cite{cohen2019certified}, which applies the neighborhood averaging and makes the neural network smoother while PTP-ADV is motivated by adversarial training, a method proposed to make the predictor capable of resisting the adversarial examples~\cite{goodfellow2014explaining} as well as boosting the clean accuracy of the predictors~\cite{xie2020adversarial, freelb}. Moreover, in order to bring more flexibility and make our exploration more comprehensive, we apply perturbations to both text~(discrete) and embedding~(continuous) space, as depicted in Figure~\ref{fig:teaser}.

\par 
% In the experiments, we test our proposed PTP algorithm and verify it can significantly reduce the training instability of prompt tuning. 
In the experiments, we conduct extensive experiments to evaluate our proposed algorithms, PTP-RN and PTP-ADV, on SuperGLUE~\cite{superglue} and FewGLUE~\cite{fewglue} benchmark. By applying the PTP algorithm on text or embedding space to the existing prompt tuning methods, we can boost the performance of prompt tuning on SuperGLUE and FewGLUE benchmarks by 1.94\% and 2.34\%  as well as make prompt tuning more stable. We also present a comprehensive ablation study and analysis of our algorithms with different perturbations. 

\par Our contributions can be summarized as:
\begin{itemize}
\setlength{\itemsep}{0pt}
\item We propose a new PTP algorithm to tackle the training instability problem in prompt tuning, which can also boost the performance. Together with PTP algorithm, we design two types of perturbations as our implicit regularizers, which are Random-Noise-based perturbation~(PTP-RN) and ADVersarial-based perturbation~(PTP-ADV). 
% Particularly, the proposed PTP-ADV is the first known effort of adversarial training on prompt tuning.
\item Moreover, as depicted in Figure~\ref{fig:teaser}, our proposed PTP-ADV and PTP-RN can be applied to both text space and embedding space, which makes our perturbations more flexible.
\item We conduct extensive experiments to evaluate the effectiveness of our algorithms on SuperGLUE and FewGLUE benchmarks. The experimental results demonstrate our proposed PTP algorithm can boost the standard performance of prompt tuning on FewGLUE and SuperGLUE by 2.34\% and 1.94\%, respectively. It also shows the great power of our algorithm in improving training stability.
% performance increase
\end{itemize}

\section{Preliminaries and Related Work} 
% formally define prompt tuning method here
% In this section, we will introduce the preliminaries and the related work of prompt tuning and adversarial training.
% \subsection{Pretained Language Models}
% Releasing the burden of training models from scratch while keeping the outstanding performance on downstream tasks, pretrained LMs brought NLP to a new era.~\cite{gpt3, liu2019roberta}. Training with gigantic unannotated corpus, like Wikipedia dumps, LMs are capable of learning not only the language written but how the various words being used.~\cite{BERT}

\subsection{Prompt Tuning}
% \paragraph{Discrete prompt tuning.} It is the method proposed by \citet{discretePT}.

\paragraph{Discrete Prompt.} 
Discrete prompt, also known as hard prompt~\cite{liu2021pre, DavisonFR19, jiang2020can, HavivBG21}, is typically a template composed of task descriptions and original input texts. \citet{gpt3} created templates for GPT-3 based on their introspection and make it suitable for various downstream tasks, such as machine translation, QA, \emph{etc}. Utilizing discrete prompts, they can achieve stunning results on NLU tasks under few-shot learning settings. By employing the predefined templates and converting the tasks into cloze questions, \citet{schick2020exploiting, fewglue} showed that even with `greener' backbone~\cite{lan2019albert}, which has 100x fewer parameters than GPT-3, they can also reach prevailing results on the few-shot version of SuperGLUE benchmark~\cite{superglue}~(also known as FewGLUE). \citet{ptuning2021} utilized the existing discrete templates and tuned embeddings of the selected tokens, which achieves SOTA results on FewGLUE. As the few-shot scenario is common and useful, in this paper, we also test the few-shot learning ability of our PTP in our experiments.

\paragraph{Continuous Prompt Tuning.} As prompts aim to boost LM's performance, it is not necessary to make tokens of prompts interpretable. Without the limit of tokens being natural words, \citet{li2021prefix} proposed to prepend a series of tunable embeddings~$E_T$~to the input embedding sequence as prompt and optimize them with training data, which provides in-context information for LMs to condition on. \citet{lester2021power}~prepended a sequence of special tokens~$T$ to the input sequence, and similarly, they tune the embeddings~$E_T$~(embeddings of special tokens) on the downstream tasks. Moreover, to further leverage the power of prompt embeddings, \citet{ptuningv2} presented PT2, a method that adds the trainable prompt embeddings to every layer of pretrained LMs as prefix embeddings. To keep the consistency of the notation, we also apply the same prompt embedding representation~$E_T$~to represent trainable prompts in every layer of LMs.

\subsection{Adversarial Training}
\paragraph{Continuous Space AT.} Over the past few~years, Adversarial Training~(AT) has demonstrated impressive results in improving model robustness~\cite{goodfellow2014explaining, tramer2017ensemble, athalye2018obfuscated}. AT can be formulated as a min-max optimization problem:
\begin{equation}
\min_\theta \mathbb{E}_{\left(x_i, y_i\right) \sim \mathbb{D}}\left[\max_{\|\delta\|_p \leq \epsilon} \mathcal{L}\left(\theta, x+\delta, y\right)\right],
\end{equation}
where $\mathcal{L}$ is the loss function, $\|\cdot\|_p$ represents $\ell_p$-norm distance and $\epsilon$ denotes the perturbation budget. \citet{madry2018PGD} proposed PGD algorithm to compute an adversarial example~(inner maximization problem) iteratively as:
\begin{equation}
\label{eq:perturbation}
    \delta^{t+1}=\Pi_{\|\delta\|_\infty \leq \epsilon}\left(\delta^{t}+\alpha \operatorname{sign}\left(\nabla_\delta L(\theta, \delta^t, y)\right)\right),
\end{equation}
where t is the iteration step; $\Pi_{\|\delta\|_\infty \leq \epsilon}$ projects perturbation $\delta$ into $\epsilon$-ball.

Besides enhancing the robustness against adversarial examples, adversarial training has been shown great power in boosting the standard performance of image classifier~\citep{xie2020adversarial}, visual language representation learning~\citep{gan2020vl}, and GNN~\citep{kong2022robust}. In this paper, we also apply a similar idea to prompt tuning and focus on boosting its performance rather than its adversarial robustness.
% Motivated by them, we explore adversarial training on boosting the standard performance of prompt tuning in this paper. % prompt tuning is not stable.

\paragraph{Discrete Space AT.} Different from adversarial attacks on images, in NLP attack, due to the discrete nature of text space, it is typically formulated as a combinatorial optimization problem to create adversarial input sequence~$s^\prime$, which is classically solved by heuristic search while maintaining the semantic similarity of the original input sequence~$s$~\cite{li-etal-2020-bert-attack, ren2019generating, morris2020textattack}. However, the searching algorithms of adversarial attacks, such as beam search~\cite{ebrahimi2017hotflip}, greedy search based on word importance~\cite{ren2019generating}, and deletion-based searching~\cite{textfooler}, are usually slow because of the high computation cost on sentence encoding and large search space~\cite{yoo2020searching}. 
% To generate adversarial examples on a 20-word sentence, if there are 10 substitution candidates for each word, the search space will be around $20^{10}$. 
% To speed up the heuristic search
\citet{a2t}~proposed A2T algorithm to accelerate the heuristic search process, which replaces the slow USE encoder~\cite{li-etal-2020-bert-attack} with DistilBERT~\cite{sanh2019distilbert} to calculate the cosine similarity between original input text and perturbed text, and they obtain significant speedup comparing to the textfooler~\cite{textfooler}. Thus, in this paper, we adapt the attacking algorithm in~\cite{a2t} to generate noises in text space.

\section{Proposed Framework}
\begin{table}[]
\small
\centering
\begin{tabular}{c|ccc}
\toprule[1pt]
     & RTE              & BoolQ            & WiC                        \\ \midrule
Acc. & $ 87.7 \pm 1.81$ & $ 83.9 \pm 0.92$ & $ 72.0 \pm 1.38$       \\ 
Var(FT) &  	0.12	& 0.09	& 0.11                    \\
Var(PT) & 1.45 & 0.74 & 1.16 \\ 
\bottomrule[1pt]

\end{tabular}
\caption{Re-implementation results of PT2~\cite{ptuningv2} with RoBERTa-large backbone on SuperGLUE benchmark. Acc.: mean accuracy. Var: variance score computed by 5 runs with different random seeds. FT: fine-tuning. PT: prompt-tuning.}
\label{tab: var. of PT2}
\end{table}

% \par In section~\ref{training objective}, we introduce the preliminary, including the final input and objective of prompt tuning.\cmh{kind weird} In section~\ref{sec:formulations}, we introduce the formulations of our proposed PTP-RN and PTP-ADV algorithms. Then we provide the implementation details on embedding~(continuous) space and on text~(discrete) space in 
% section~\ref{subsec: embedding space perturbation}~and~\ref{subsec: text space perturbation}. Finally, we provide the description of our whole training framework in section~\ref{sec:framework}.

% \cmh{I don't see the motivation on training stability and some explanation in Figure 2}
% \cmh{still think the methodology part is not good. It is more like listing instead of persuading people why you need to do that and why it helps and works}

\subsection{Preliminary}
\label{training objective}

Before introducing the proposed algorithms, we first briefly describe the word to embedding process of LMs as well as the final embedding input of the continuous prompt tuning. Given the $n$ word input sequence~$s=\left[s_1, \ldots, s_n\right]$ and word to embedding function~$f_V$, where $V$ denotes embedding matrix, the input embedding~$E_s=\left[e\left(s_1\right), \ldots, e\left(s_n\right)\right] \in \mathbb{R}^{n \times d}$~can be obtained by~$E_s=f_{V}(s)$, where $d$ denotes the dimension of the word embedding. Let continuous prompts be~$E_T=[e_t^1, \ldots, e_t^m] \in \mathbb{R}^{m \times d}$, the update of~$E_T$~can be either directly, or through a reparameterization encoder~$P$ like MLP/LSTM,~$P(E_T)=[h_0, \ldots, h_m] \in \mathbb{R}^{m \times d}$. For simplicity, we still use~$E_T$~to represent the output of~$P(E_T)$, and the final input embedding sequence is written as~$\left[E_T; E_s\right] \in \mathbb{R}^{(m+n) \times d}$. 
\par The training objective of continuous prompt tuning can be formulated as:
\begin{equation}
\min_{\theta} \mathbb{E}_{\left(s, y\right) \sim \mathbb{D}}\left[ \mathcal{L}\left(\mathcal{M}\left(\theta, s, y\right)\right)\right],
\label{eq:spt}
\end{equation}
where $M$ denotes LM; $\theta$ represents the trainable parameters of $E_T$ and prompt encoder $P$ while $\mathbb{D}$ is the underlying data distribution.

\SetKwComment{Comment}{/* }{ */}
\RestyleAlgo{ruled}

% PGD attacks on propmt tuning
\begin{algorithm}[t]
\caption{PGD on prompt tuning}
\label{algorithm: PGD}
\textbf{Require:} Perturbation iteration~$n$ and size~$\alpha$. The bound of perturbation~$\epsilon$ \;
\For{epoch = $1, \ldots, n$}{
$E_s$.requires\_grad $\gets$~True\;
$\hat{y_i}\gets \arg \max_{y}\left(\Pr\left[y|\mathcal{M}\left(I^\prime \right)\right]\right)$\;
$\mathcal{L}(\hat{y}, y)$.backward()\;
$E_s^{\prime} \gets E_s + \alpha * E_s.grad.sign()$ \;
$\delta = \Pi_{\|\delta\|_\infty \leq \epsilon}(E_s^{\prime} - E_s )$ \;
$E_s^{\prime} \gets E_s + \delta$ \;
$\mathcal{M}$.zero\_grad()\;
}
return $E_s^{\prime}$
\end{algorithm}

\begin{algorithm}[t]
\textbf{Require:} Prompt embeddings $E_T$; input embeddings~$E_s$; trainable parameter~$\theta$ for prompt encoder~$P$ and~$E_T$; Training data $D$; Pre-trained LM~$\mathcal{M}$; Loss function $\mathcal{L}$ \;
Initialize parameters $\theta$\;
\For{epoch = $1, \ldots, K$}{
\textcolor{blue}{/* \textit{standard prompt tuning} */} \\
Sample a minibatch data~$(s,y)$ from $D$\;
$\Theta$.requires\_grad~$\gets$~True\;
$I \gets [E_T; E_s]$ \;
$\hat{y} \gets arg\max_{y}\left(\Pr\left[y|\mathcal{M}\left(I\right)\right]\right)$\;
$\mathcal{L}(\hat{y}, y)$.backward() and update $E_T$ \;
\textcolor{blue}{/* \textit{training with perturbed data} */} \\
$\Theta$.requires\_grad~$\gets$~False\;
Apply PTP-RN or PTP-ADV to $s$ or $E_s$\;
% $E_s^\prime \gets A(E_s)$ or $s^\prime = A(s)$\;
$I^\prime \gets [E_T; E_s^\prime]$ \;
$\Theta$.requires\_grad~$\gets$~True\;
$\hat{y_i}\gets arg\max_{y}\left(\Pr\left[y|\mathcal{M}\left(I^\prime \right)\right]\right)$\;
$\mathcal{L}(\hat{y}, y)$.backward() and update $E_T$\;
}
\caption{PTP}
\label{algorithm:APT}
\end{algorithm}

\subsection{Proposed Formulation}
Although continuous prompt tuning~\cite{ptuningv2} could achieve comparable performance with the fine-tuning method by only using  0.1\% to 3\% trainable parameters, it suffers from unstable training issues: as shown in Table~\ref{tab: var. of PT2}, even only changing the random seed in different runs, the final performance is very unstable.
% \cmh{will few-shot be more unstable?} \clc{just a little bit more}
To investigate the issue, we plot the loss landscape of the vanilla prompt tuning as Figure~\ref{fig:loss landscape} and observe sharp crests in a small local region with a small noise, which means a small perturbation on embedding space would cause a significant reduction in the final accuracy. It is also known as the training instability problem~\cite{chen20perturb}. To address this challenge in prompt tuning, we propose perturbation-based regularizers to force the loss landscape to be smooth.
% \label{sec:formulations}
% The training objective of standard prompt tuning is presented as Eq.~(\ref{eq:spt}). 
% \cmh{move to the beginning of the proposed method as the motivation. Maybe have some tables to show the instability problem}
% \par Though the vanilla prompt tuning can achieve superb results on NLU tasks~\cite{ptuning2021, ptuningv2}, their training is unstable and their loss landscapes are precipitous.
% One intuitive thought is to force the loss landscape of $\mathcal{L}\left(\mathcal{M}\left(\theta, s, y\right)\right)$ to be smooth. Thus, we introduce perturbation-based regularizers into it. In this way, given the perturbation~$\delta$, the two versions of perturbation-based regularizers can be formulated as follows:
Specifically, we introduce two versions of perturbation-based regularizers that can be formulated as follows:
\begin{equation}
\begin{array}{l}
\min_\theta \mathbb{E}_{\left(s, y\right) \sim \mathbb{D}} \left[ \mathcal{L}\left(\mathcal{M}\left(\theta, s+\delta, y\right)\right)\right], \text{s.t.} \\
\text{PTP-RN: } \delta \sim \mathcal{N} \\
\text{PTP-ADV: }\delta = \max_{\| \delta \| \leq \epsilon} \mathcal{L}\left(\theta, s + \delta, y\right),
\end{array}
\end{equation} 
where~$\mathcal{N}$ denotes Gaussian distribution. For PTP-RN, we minimize~$\theta$~under small random perturbation, aiming to force the model to focus on perturbed pair $(s+\delta, y)$ and have a robust prediction within the neighborhood of $s$. It is related to the idea of randomized smoothing~\cite{cohen2019certified}, which obtains a smoother predictor via randomly averaging the neighborhood of the given function. For PTP-ADV, the perturbation~$\delta$~is generated by adversarial attack algorithms such as PGD~\cite{madry2018PGD}, A2T~\cite{a2t}, and the worst-case training loss is minimized under small perturbation bounded by~$\epsilon$. The idea is motivated by adversarial training, which is usually applied as a form of adversarial defense.~\cite{goodfellow2014explaining, sprout} 
% \cmh{check how xiangning describe their method.} 

Since we are the first to investigate the training stability issue of the prompt tuning and it is still unknown which space to inject perturbation~$\delta$ is better, we apply it on both text and embedding space to bring more flexibility and make our exploration more comprehensive. The perturbed sequence~$s^\prime$ or the perturbed embedding~$E_s^\prime$ can be obtained as 
% \cmh{any thoughts on both continuous and discrete space?} 
\begin{equation}
\begin{array}{c}
    s^\prime = s + \delta, \\
    E_s^\prime = E_s + \delta,
\end{array}
\end{equation}
where~$s, E_s$~denote the input sequence and the input embedding, respectively. It is worth noticing that if the perturbation is on~$s$~(text space), through~$f_V$, the perturbed~$s^\prime$~will be converted into input embedding, which is also denoted as~$E_s^\prime$.
\par The main idea of our proposed formulation is that we force our algorithm to not only learn from the clean data pair~$(s, y)$~but also perturbed data pair~$(s+\delta, y)$~to make the training more smooth.

% We elaborate the perturbation algorithms~(Line 12, Algorithm~\ref{algorithm:APT}) in the following subsections.

\subsection{PTP-RN} 
\label{subsec: embedding space perturbation}
% Since our goal is not on creating label-preserving examples which maintain the   characteristic of LMs, converting the discrete text to continuous embeddings, we design our PTP-RN on both text space and embedding space to enhance the flexibility. 
\subsubsection{Embedding Space~(RG Perturbation)}
In embedding space, how to create perturbed examples is still an unsolved problem. But since the ultimate effects of PTP-RN are the only thing we care about, not the interpretability, it is feasible for us to add random-noise-based perturbation on word embeddings. Given the embeddings of the input sequence~$E_s = \left[e(s_0), e(s_1), \ldots, e(s_n)\right]$, where~$e(s_i) \in \mathbb{R}^d$, PTP-RN in embedding space samples~$\delta$ from Gaussian distribution and randomly selects some embeddings to perturb, which make sure. The perturbation~$\delta$~can be formulated as:
\begin{equation}
\begin{array}{c}
E_s^\prime = E_s + \delta, \text{s.t.} \\ 
\delta = \{\delta_1, 0, \ldots, \delta_i, 0\},
\end{array}
\label{eq:rn}
\end{equation}
% make delta clear
where $\delta \in \mathbb{R}^{n \times d}$ has the same length as the input embeddings; $i$~denotes the number of embeddings being perturbed and~$\delta_{n=1, \ldots, i} \sim \mathcal{N}(0, \sigma\mathbb{I}_{d})$, with~$d$ denoting the dimension of the word embedding and $\sigma$ controlling the magnitude of perturbation. We represent PTP-RN on embedding space as PTP+RG.
\subsubsection{Text Space~(RM Perturbation)}
In text space, similarly, our goal is to create label-preserving and perturbed input data to augment the training data and make the training stable. Given an input sequence $s$, PTP-RN randomly selects some tokens and converts them into [MASK]. The perturbed sequence~$s^\prime$~can be formulated as:
\begin{equation}
s^\prime = \{s_0, \left[\mathrm{MASK}\right], \ldots, \left[\mathrm{MASK}\right], s_n\},
\label{eq:RM}
\end{equation}
where we perturb~$i$~tokens. It should be noticed that unlike BERT pretraining process~\cite{BERT}, where the model predicts the label of [MASK], our model will not predict anything on the tokens we mask and we just use [MASK] token as a perturbation on discrete space. PTP+RM is used to denote PTP-RN on text embedding space.

% After the perturbation finished, we employ the perturbed~$s^\prime$ with unchanged label~$y$ to optimize the trainable parameters in~$E_T$ and~$P$, as Line 14-16 in Algorithm~\ref{algorithm:APT}.

\begin{table*}[t]
\centering
\begin{tabular}{c|cc|cc|cc|cc}
\toprule[1.5pt]
Method  & \multicolumn{2}{c|}{BoolQ} & \multicolumn{2}{c|}{CB} & \multicolumn{2}{c|}{WiC} & \multicolumn{2}{c}{RTE} \\ \cline{2-9} 
        & BERT       & RoBERTa       & BERT      & RoBERTa     & BERT      & RoBERTa      & BERT      & RoBERTa     \\ \hline
FT      & 77.7       & 86.9          &94.6       &98.2        &74.9      &75.6            &70.4     & 86.6          \\
PT2     & 75.8       & 84.8          &94.6       &100         &75.1      &73.4            &78.3     & 89.5           \\
PTP+A2T  & 76.4      & 85.7          & 94.5      &  99.6     & 75.8        & 72.9         & 78.6   & 89.7            \\
PTP+RG & 77.3       & 86.2        &95.8      &  99.8        & \textbf{76.7}\hld{$\uparrow$1.6} & 75.5         &  79.9  & 90.6                \\
PTP+RM  & 77.4       &  85.9      & 95.7         & 100      & 76.4       & 75.2     & 79.6    &  91.6           \\
PTP+PGD & \textbf{78.3}\hld{$\uparrow$2.5} &  \textbf{86.7}\hld{$\uparrow$1.9} &\textbf{96.1}\hld{$\uparrow$1.5}  & \textbf{100}\hld{$\uparrow$0}  & 76.6 & \textbf{75.7}\hld{$\uparrow$2.3} & \textbf{80.3}\hld{$\uparrow$2.0} & \textbf{92.0}\hld{$\uparrow$2.5} \\ 
\bottomrule[1pt]
\end{tabular}
\begin{tabular}{c|cc|cc|cc|cc}
\toprule[1pt]
Method  & \multicolumn{2}{c|}{COPA} & \multicolumn{2}{c|}{MultiRC(F1a)} & \multicolumn{2}{c|}{ReCoRD} & \multicolumn{2}{c}{WSC} \\ \cline{2-9} 
        & BERT       & RoBERTa       & BERT      & RoBERTa     & BERT      & RoBERTa      & BERT      & RoBERTa     \\ \hline
FT      & 69.0       &  94.0        & 70.5       & 85.7       & 70.6       & 89.0          & 68.3      & 63.5       \\
PT2     & 73.0       &  93.0        & 70.6       & 82.5       & 72.8       & 89.3          & 68.3      & 63.5       \\
PTP+A2T & 73.3       &  93.2        & 71.4       & 82.6       & 73.6       & 89.7        & 68.5       & 63.8            \\
PTP+RG  &\textbf{75.1}\hld{$\uparrow$2.1} &  93.9      & 72.6    & \textbf{84.9}\hld{$\uparrow$2.4} & 74.9       & 90.5        & 69.4        & 65.0             \\
PTP+RM  &74.6          & 93.8       & 72.9      & 84.4      & 74.8         & 90.6         & 69.2        & 64.8            \\
PTP+PGD & 74.7  & \textbf{94.1}\hld{$\uparrow$1.1}     & \textbf{73.4}\hld{$\uparrow$2.8} & 84.6 & \textbf{75.1}\hld{$\uparrow$2.1} & \textbf{91.9}\hld{$\uparrow$1.6} & \textbf{69.7}\hld{$\uparrow$1.4} & \textbf{65.0}\hld{$\uparrow$1.5}  \\ 
\bottomrule[1.5pt]
\end{tabular}
\caption{Results of our proposed PTP algorithm in fully-supervised learning settings. We employ the large-size version of BERT and RoBERTa models~(BERT-Large size: 335M and RoBERTa-large size: 355M, respectively). We use bold font to mark the best and red subscript to mark the improvement compared to the PT2.}
\label{tab: fully-supervised}
\end{table*}

\begin{table*}[ht]
\centering
\begin{tabular}{c|c|c|c|c|cc|c|c}
\toprule[1.5pt]
(Dev 32)   & BoolQ  &CB           & WiC    & RTE    & \multicolumn{2}{c|}{MultiRC}      & WSC    & COPA   \\
Method & (Acc.) & (F1) & (Acc.) & (Acc.) & \multicolumn{1}{c|}{(EM)} & (F1a) & (Acc.) & (Acc.) \\ \hline
PET Best     &  75.1  & 83.5 & 52.6 & 65.7 & \multicolumn{1}{c|}{35.2} & 75.0  & 80.4 & 83.3   \\
% Best PET &  75.1 & \multicolumn{1}{c|}{86.9}   &   83.5  &   52.6 &  65.7 & \multicolumn{1}{c|}{35.2}  &  74.5 & 80.4 & 83.3      \\
PT       &  77.8   &   92.3  &   56.3 &  76.5 &\multicolumn{1}{c|}{36.1}  & 75.0 &   84.6 & 87.0     \\
PTP+RM   &   79.9   & 93.2 & 58.1  & \textbf{78.6}\hld{$\uparrow$2.1}   & \multicolumn{1}{c|}{36.2} & 78.3 & 85.9 &88.6 \\
PTP+RG    &  79.5  &  \textbf{93.7}\hld{$\uparrow$1.4} &  58.0 & 77.7  & \multicolumn{1}{c|}{36.6}  & 78.1  & 85.4 & 88.3 \\
PTP+A2T   &  78.6  & 92.6  & 56.6 & 77.4  & \multicolumn{1}{c|}{36.5} & 76.0  & 84.7  &  87.7  \\ 
PTP+PGD   &  \textbf{80.2} \hld{$\uparrow$2.4} &  93.5 & \textbf{58.5} \hld{$\uparrow$2.2} & 78.5  & \multicolumn{1}{c|}{\textbf{37.4}\hld{$\uparrow$1.3}}  & \textbf{78.9}\hld{$\uparrow$3.9}& \textbf{86.0}\hld{$\uparrow$1.4} & \textbf{88.9}\hld{$\uparrow$1.9} \\ \hline
PET(Full Dev)  &  79.4   &   59.4 &  52.4 &\multicolumn{1}{c|}{69.8}  & 37.9 &  77.3 & 80.1 & 95.0  \\
iPET(Full Dev) &  80.6   &   92.4  &   52.2 &  74.0 &\multicolumn{1}{c|}{33.0}  & 74.0 &   - & -  \\
\bottomrule[1.5pt]
\end{tabular}
\caption{Results of our PTP algorithm in Few-shot learning(32 training examples) settings. PT: P-tuning~\cite{ptuning2021} and the backbone LM is alberta-xxl-v2. We use \textbf{bold} font to mark the best. The red subscript denotes the increase of our method compared with the baseline method PT. Dev 32: development set contains 32 unused examples from the training set, same as~\cite{ptuning2021}. Full Dev: original development set.}
\label{Tab: Fewshot}
\end{table*}

\subsection{PTP-ADV}
\label{subsec: text space perturbation}
\subsubsection{Embedding Space~(PGD Perturbation)}
Different from previous PGD training methods which focus on improving the models' robustness, we aim to smooth the loss landscapes and boost the performance of prompt tuning by adding adversarial-based regularization. Given the embedding sequence~$E_s$, PTP-ADV adopts multi-step PGD to generate perturbations on embedding space. The perturbation~$\delta$ is computed iteratively as:
% the aim of PGD algorithm is to make the LMs predict the label falsely. Different from the PGD attacks in images, where all the pixels can be added the gradient-based perturbations and there is no trainable parameters in the input, our input contains trainable prompt embeddings~$E_T$. Therefore, we are supposed to fix the trainable parameters~$\theta$~in $E_T$~and prompt encoder~$P$, then we can have the perturbation update in APT with PGD attack according to Eq.~\ref{eq:perturbation}:
\begin{equation}
\begin{array}{c}
E_s^\prime = E_s + \delta^t, \text{s.t.}\\
\delta^{t} =\Pi_{\|\delta\|_\infty \leq \epsilon}\left(\delta^{t-1}+\alpha \nabla_\delta L \right)
% U^t = \operatorname{sign}\left(\nabla_\delta L(\theta, \mathcal{M}([E_T; E_s+\delta^t]), y)\right),
\end{array}
\label{eq:pt-pgd}
\end{equation} 
where~$\delta^t$~ denotes the~$t$-th iterations of PGD perturbation and it will be added to the input embedding sequence~$E_s$ after all the iterations are finished. Algorithm~\ref{algorithm: PGD} shows the implementation details of our PGD attack on prompt tuning. PTP+PGD is applied to denote our PTP-ADV algorithm with perturbation on embedding space.

\subsubsection{Text Space~(A2T Perturbation)}
Furthermore, to enhance the flexibility of PTP-ADV and boost model generalization ability, we apply it to the text space: PTP-ADV adopts the attack algorithm in A2T~\cite{a2t} to generate its perturbation~$\delta$, which is an algorithm composed of NLP attack and adversarial training. Given the input sequence~$s$, the perturbed sequence~$s^\prime$, with A2T perturbation, can be represented as:
\begin{equation}
    s^\prime = \{s_0, s^\prime_1, \ldots, s^\prime_{n-1}, s_n\},
\end{equation}
where $s^\prime_i$ denotes the perturbed word. For simplicity, we also call it PTP+A2T.

%training with perturbed data
% \subsection{Description of Framework} 
% \label{sec:framework}
Algorithm~\ref{algorithm:APT} provides the details about the whole training process of PTP algorithm. In the standard prompt tuning, the input of LM is composed of prompt embedding~$E_T$~and input embedding~$E_s$, which is denoted as $I = [E_T;E_s]$. After LM gives a prediction of~$I$, we backpropagate the loss to update~$E_T$. In training with perturbed data part~(Line 10-16, Algorithm~\ref{algorithm:APT}), the discrete or continuous space perturbations of input data are generated by PTP-RN or PTP-ADV firstly. Then the perturbed input is employed to conduct training with original label~$y$, which also plays a data-augmentation role to boost the performance of the prompt tuning. 

% We freeze all the trainable parameters~$\Theta$ while generating perturbations~(Line 11, Algorithm~\ref{algorithm:APT}). 
% The biggest difference between ours and previous prompt tuning is that we conduct one addition training with perturbed data step in each epoch. 

\section{Experiments}
We conducted empirical studies on two popular natural language understanding~(NLU) benchmarks: SuperGLUE benchmark~\cite{superglue} and FewGLUE benchmark~\cite{fewglue}. We tested the proposed framework in both fully-supervised and few-shot settings to verify the effectiveness of our proposed PTP-RN and PTP-ADV algorithm with perturbations on both text and embedding space.  
% \cmh{IMPORTANT: I think we haven't included the definition of fully-supervised and few-shot settings? }
% Experiments are all conducted on servers with RTX A6000 GPUs, each of them having 48GB memory. 
% Tasks:
% Specifically, MultiRC~\cite{multirc} and BoolQ~\cite{clark2019boolq} are question answering tasks. \cmh{?} RTE~\cite{dagan2005RTE} and CB~\cite{de2019CB} are tasks aiming at recognizing textual entailment. ReCoRD~\cite{zhang2018record} is a multiple-choice QA task evaluating the reading comprehension ability. Cast as binary classification of sentence pairs, the goal of WiC~\cite{pilehvar2018wic} is to disambiguate word sense. WSC~\cite{levesque2012WSC} is a conference resolution task and COPA~\cite{gordon2012copa} is a causal reasoning task. 

% \begin{figure*}[ht]
%     \centering
%     \includegraphics[width=0.98\textwidth]{figures/fsl.pdf}
%     \caption{Performance of PTP+PGD on FewGLUE~(WiC, BoolQ, WSC datasets) with different $\alpha$ and PGD iterations. The dashed red line represents the performance of the baseline method PT~\cite{ptuning2021}. It shows the best~$\alpha$ and PGD iterations are 1e-3 and 4, respectively. \clc{TODO: remove the grey background}}
%     \label{fig:few-pgd.}
% \end{figure*}

\subsection{Experimental Setup}

\paragraph{NLU Dataset.} SuperGLUE benchmark~\cite{superglue} contains 8 challenging natural language understanding~(NLU) tasks. We also include the few-shot version of SuperGLUE, FewGLUE benchmark~\cite{fewglue} to test the ability of our algorithm, which consists of 32 training samples in each dataset on SuperGLUE. Following~\cite{fewglue, ptuning2021}, we report results on 7 of 8 NLU tasks in few-shot settings. 

\paragraph{Fully \& Few-shot Setting.} In fully-supervised setting, the full training set of each task in SuperGLUE~\cite{superglue} is employed during the prompt tuning process. Besides, in the model selection part, we adopt the whole validation set. As few-shot learning ability of prompt tuning can reduce the cost of annotations in real-world applications, following~\cite{fewglue, ptuning2021}, we also test our algorithm under few-shot settings. To be specific, we use the training set provided by FewGLUE~\cite{fewglue}, the few-shot version of SuperGLUE, containing 32 training pairs in each task. Besides, we use the same version of the development set as~\cite{ptuning2021} to select models, which are created by randomly choosing 32 unused training pairs.

\paragraph{Baseline Methods.} We include 2 prompt tuning methods P-tuning~\cite{ptuning2021} (PT) and P-tuning-v2~\cite{ptuningv2} (PT2) as baselines. PT is the state-of-the-art method in FewGLUE benchmark while PT2 also achieves excellent performance in SuperGLUE benchmark. We defer the hyperparameters such as learning rate and prompt length in Appendix~\ref{sec: details of lr and pl}. We also leave some figures and tables in Appendix.

\paragraph{Pretrained LMs.} Following the settings in~\cite{ptuningv2, ptuning2021}, we include BERT-large~\cite{devlin2018bert} and RoBERTa-large~\cite{liu2019roberta} for fully-supervised settings and ALBERTA-xxlarge-v2~\cite{lan2019albert} for few-shot settings. To have a fair comparison with the baseline methods, for fully-supervised settings, all backbone LMs are frozen, except in fine-tuning, same as~\cite{ptuningv2}. For few-shot learning settings, backbone LMs are tuned with trainable prompt embeddings, same as~\cite{ptuning2021}.

% \paragraph{Prompt Length and Prompt Learning Rate.} We defer the learning rate and prompt length details to Appendix~\ref{sec: details of lr and pl}.

\subsection{Results on Fully-supervised Setting}
In fully-supervised settings, Table~\ref{tab: fully-supervised} demonstrates the results of our proposed PTP algorithm with 4 different perturbations on all 8 tasks of SuperGLUE benchmark. It is worth noticing that PTP+PGD achieves the best performance in almost all datasets except WiC~(BERT), COPA(BERT), and MultiRC~(RoBERTa). Overall, the best method PTP+PGD outperforms the baseline method PT2 by 1.94\%~(with BERT-large backbone) and 1.63\%~(with RoBERTa-large backbone) on average.

\paragraph{Text vs. Embedding Perturbation.} PTP with PGD and RG perturbation on continuous space~(embedding space) are perform better than PTP with RM and A2T, which indicates perturbing on continuous space is more effective than perturbing on discrete space in fully-supervised settings. As for pretrained LMs~(BERT-large and RoBERTa-large), results show the superb learning ability of our PTP algorithm regardless of which backbone.

\subsection{Results on Few-shot Setting}
In few-shot learning settings, we employ FewGLUE, also known as few-shot version of SuperGLUE. PET~\cite{fewglue} and iPET~\cite{schick2020exploiting} are the methods using discrete prompts. We test the previous SOTA method on FewGLUE, PT~\cite{ptuning2021}~, as our baseline method and validate on the same development set~(Dev 32). As illustrated in~\cite{ptuning2021}, for a fair comparison, the results of PET Besst~(Dev 32) are reported as removing all the additional tricks like ensemble, distillation, etc. PET~(Full Dev) and iPET~(Full Dev) denote the methods with the original validation set.
\par Our main results are shown in Table~\ref{Tab: Fewshot}. PTP achieves better results than the previous state-of-the-art method PT in all 7 tasks, which verifies the effectiveness of our algorithms in few-shot NLU tasks. Especially, PTP+PGD outperforms the previous PT by 2.34\% on average. Comparing the PTP+PGD~(Dev 32) to the methods with the original dev set, it still has an advantage on most of the tasks~(5 of 7) while the results are similar in BoolQ~(better than PET with full dev set but worse than iPET). The PTP with RM and RG perturbation method also achieve remarkable improvement when compared to the baseline method PT. Moreover, PTP with A2T perturbation can also boost the performance of the baseline by a small margin.

\subsection{Results on Improving Training Stability} 
\begin{figure}[t]
    \centering
    \includegraphics[width=0.45\textwidth]{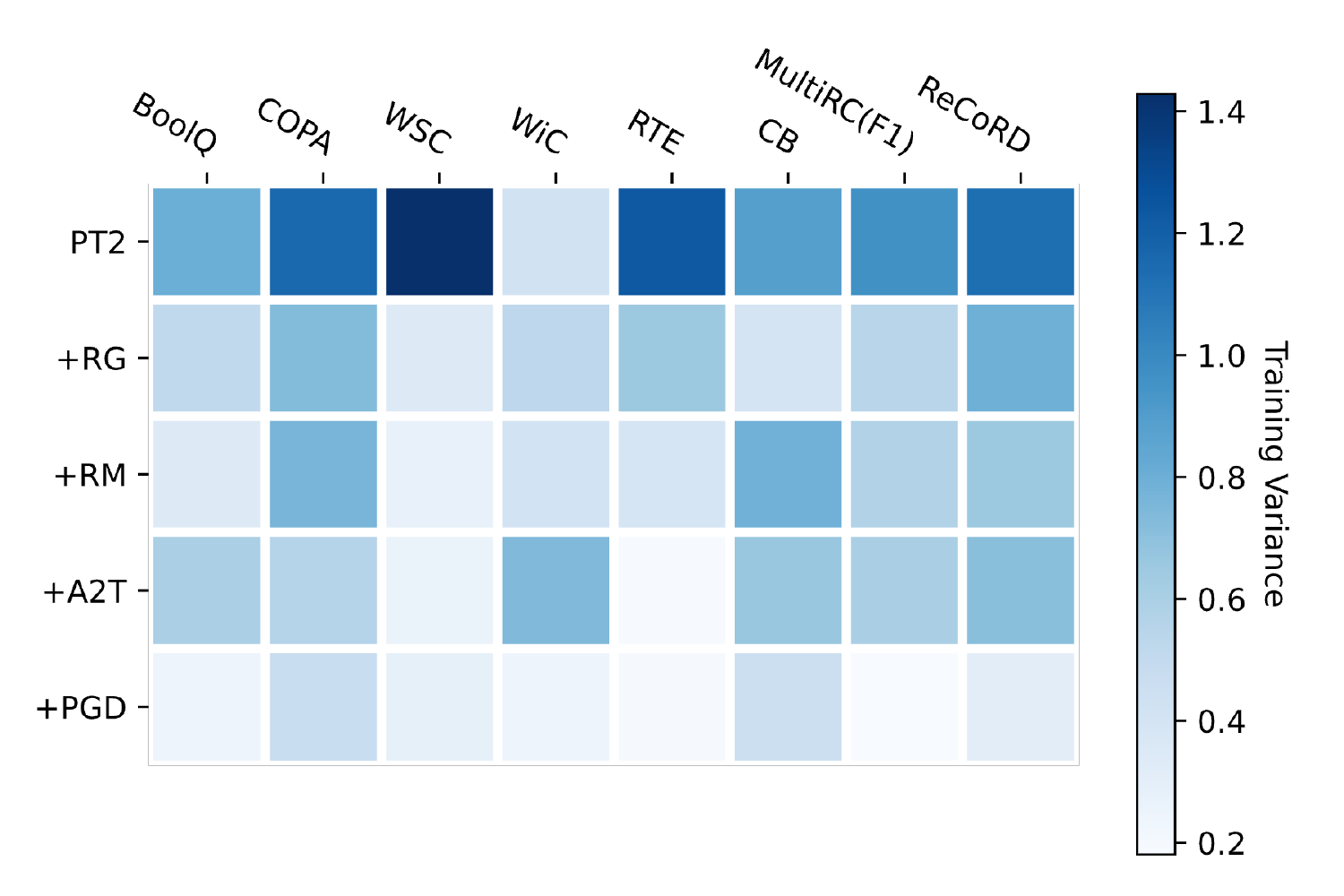}
    \caption{The variance of the scores on the dev sets from SuperGLUE Benchmark. We compute it on 5 runs with different random seeds to report. The reported experiments are all using BERT-large models as backbones. }
    \label{fig:deviation.}
\end{figure}

In addition to improved generalization performance, the proposed method could stable the training process. Figure~\ref{fig:loss landscape} provides strong evidence that our proposed PTP-RN and PTP-ADV training methods have a much smoother loss landscape compared to the vanilla prompt tuning. For the few-shot learning setting, we demonstrate the results in Table~\ref{tab:robustness}. It could be seen that all our PTP methods have smaller variances than the baseline method. Specifically, PTP+PGD has the smallest variance in 5 runs, which indicates its training and generalization stability. Compared with the PTP-RN methods~(RG, RM), PTP-ADV methods~(A2T, PGD) achieve smaller variance. We also conduct the experiments under fully-supervised learning settings in Figure~\ref{fig:deviation.}. It shows that in all 8 tasks from SuperGLUE benchmark, our proposed method can still reduce the training variance of different runs with the same hyperparameter except the seeds.

% \begin{figure*}[t]
%     \centering
%     \includegraphics[width=0.99\textwidth]{figures/full-rn.pdf}
%     \caption{Performance of PTP+RG on SuperGLUE~(WiC, RTE, COPA datasets) with $\sigma$ from \{1e-2, 1e-3, 1e-4\} and perturbed embeddings from \{1, 5, 10, 20\}. The dashed red line represents the performance of baseline method PT2 with BERT-large as backbone LM. }
%     \label{fig:few-rn.}
% \end{figure*}

% Ablation Study of our paper
\section{Ablation Study} % focus on the hyperparameter choosing of different settings
In this section, we conduct comprehensive ablation studies of our perturbation methods under both fully-supervised and few-shot learning settings. 

\paragraph{RG Perturbation.}
We investigate the strength of gaussian noise~$\delta$~and number of perturbed embeddings~$i$~affects the performance~(see Eq.~(\ref{eq:rn})). Specifically, we run experiments with~$\sigma$~, which is the variance of the added Gaussian noise, from \{$10^{-4}$, $10^{-3}$, $10^{-2}$\} and the number of word embeddings perturbed, which is denoted as~$i$~in Eq.~(\ref{eq:rn}), from \{1, 5, 10, 20\}. Under the fully-supervised learning setting, we report the results on COPA, RTE, WiC tasks in Figure~\ref{fig:few-rn.}. The results show the appropriate~$\sigma$~is supposed to be $10^{-2}$ and the number of perturbed embeddings to be 5. With large~$\sigma$ and large~$i$, PTP+RG is more likely to fail in comparison to the baseline method. Under few-shot learning settings, we select results in MultiRC task to report, as shown in Table~\ref{tab: few-MultiRC}. The encouraging result also demonstrates that the best choice of~$\sigma$~and number of embeddings perturbed is $10^{-3}$ and 5, respectively. 

\paragraph{PGD Perturbation.}
We investigate how different~$\alpha$~(See Eq.~(\ref{eq:pt-pgd})) and PGD iterations affect the performance.
We run experiments with~$\alpha$~from \{$10^{-4}$, $10^{-3}$, $10^{-2}$\} and PGD iterations from 1 to 5. Under fully-supervised learning settings, we present the the results of COPA dataset in Table~\ref{tab:full-pgd}. It shows that large~$\alpha$~in PGD will be detrimental to the performance. Under few-shot learning settings, Figure~\ref{fig:few-pgd.} demonstrates the results of different~$\alpha$ and different iterations of PTP+PGD on few-shot settings. In all 3 datasets, when~$\alpha$ is $10^{-3}$, not too small nor too large, and PGD iters is~$4$, PTP+PGD can achieve outstanding performance.

\paragraph{RM Perturbation.} 
We investigate how different numbers of random [MASK] inserted affects the PTP. Formally, the number of [MASK] inserted is defined as~$i$ in Eq.~(\ref{eq:RM}). We conduct experiments with~$i$ from 1 to 10. Under few-shot learning settings, Figure~\ref{fig:few-shot setting PT+RM.}~presents the results of PTP with RM perturbation on FewGLUE~(MultiRC and RTE dataset). We observe that RM perturbation can boost the performance substantially in few-shot settings and the best choice of $i$ is 8. Under fully-supervised settings, Table~\ref{tab:ablationRM} presents the ablation of RM perturbation. It also shows that a large number of [MASK] inserted harms the performance, especially when the backbone is RoBERTa.

% table for A2T ablation
% \begin{table}[t]
% \begin{tabular}{c|cccc}
% \toprule[1pt]
% Cosine Sim    & 0.2  & 0.4  & 0.6  & 0.8  \\ \hline
% BoolQ(Full)   & -0.83 & -0.46 & +0.61 & +0.52 \\
% BoolQ(Few)    &  -0.96 &   -0.13   &  +0.78   & +0.84     \\
% MultiRC(Full) &  -0.77 &   -0.35   &   +0.25   & +0.71     \\
% MultiRC(Few)  &   -0.59  &  -0.54    &   +1.03   & +0.68    \\
% \bottomrule[1pt]
% \end{tabular}
% \caption{The results of different minimum cosine similarity in A2T perturbation. Full: fully-supervised learning setting. Few: few-shot learning setting.}
% \label{tab:A2T ablation}
% \end{table}

\paragraph{A2T Perturbation.}
We investigate how the minimum cosine similarity between normal input~$s$ and perturbed input~$s^\prime$ of A2T perturbation affects the results. We run experiments with minimum cosine similarity from \{0.2, 0.4, 0.6, 0.8\} and show results in Table~\ref{tab:A2T ablation}. It indicates small similarity may cause damage to the standard performance because the perturbation can be too large in this case.

\section{Conclusion}
\par In this paper, we first investigated the training instability issues on prompt tuning, which has a precipitous loss landscape in its visualization. To tackle the problem, we proposed PTP-RN and PTP-ADV algorithms, which include four different perturbations~(RG, RM, ADV, A2T) on both discrete and continuous spaces, to smooth the loss landscape and make the training stable. Furthermore, our algorithms are also capable of boosting the performance of prompt tuning. The extensive experiments validate the effectiveness of our proposed algorithms on NLU tasks under both fully-supervised and few-shot settings. 

% \section*{Ethical Concern}
% We utilize adversarial attacks algorithms for adversarial training on prompt tuning. Though adversarial attacks are detrimental to the LMs, we demonstrate we can improve the performance of prompt tuning. 

\bibliography{anthology,custom}

\begin{thebibliography}{44}
\expandafter\ifx\csname natexlab\endcsname\relax\def\natexlab#1{#1}\fi

\bibitem[{Athalye et~al.(2018)Athalye, Carlini, and
  Wagner}]{athalye2018obfuscated}
Anish Athalye, Nicholas Carlini, and David~A. Wagner. 2018.
\newblock \href {http://proceedings.mlr.press/v80/athalye18a.html} {Obfuscated
  gradients give a false sense of security: Circumventing defenses to
  adversarial examples}.
\newblock In \emph{Proceedings of the 35th International Conference on Machine
  Learning, {ICML} 2018, Stockholmsm{\"{a}}ssan, Stockholm, Sweden, July 10-15,
  2018}, volume~80 of \emph{Proceedings of Machine Learning Research}, pages
  274--283. {PMLR}.

\bibitem[{Brown et~al.(2020)Brown, Mann, Ryder, Subbiah, Kaplan, Dhariwal,
  Neelakantan, Shyam, Sastry, Askell, Agarwal, Herbert-Voss, Krueger, Henighan,
  Child, Ramesh, Ziegler, Wu, Winter, Hesse, Chen, Sigler, Litwin, Gray, Chess,
  Clark, Berner, McCandlish, Radford, Sutskever, and Amodei}]{gpt3}
Tom Brown, Benjamin Mann, Nick Ryder, Melanie Subbiah, Jared~D Kaplan, Prafulla
  Dhariwal, Arvind Neelakantan, Pranav Shyam, Girish Sastry, Amanda Askell,
  Sandhini Agarwal, Ariel Herbert-Voss, Gretchen Krueger, Tom Henighan, Rewon
  Child, Aditya Ramesh, Daniel Ziegler, Jeffrey Wu, Clemens Winter, Chris
  Hesse, Mark Chen, Eric Sigler, Mateusz Litwin, Scott Gray, Benjamin Chess,
  Jack Clark, Christopher Berner, Sam McCandlish, Alec Radford, Ilya Sutskever,
  and Dario Amodei. 2020.
\newblock \href
  {https://proceedings.neurips.cc/paper/2020/file/1457c0d6bfcb4967418bfb8ac142f64a-Paper.pdf}
  {Language models are few-shot learners}.
\newblock In \emph{Advances in Neural Information Processing Systems},
  volume~33, pages 1877--1901.

\bibitem[{Chen and Hsieh(2020)}]{chen20perturb}
Xiangning Chen and Cho{-}Jui Hsieh. 2020.
\newblock \href {http://proceedings.mlr.press/v119/chen20f.html} {Stabilizing
  differentiable architecture search via perturbation-based regularization}.
\newblock In \emph{Proceedings of the 37th International Conference on Machine
  Learning, {ICML} 2020, 13-18 July 2020, Virtual Event}, volume 119 of
  \emph{Proceedings of Machine Learning Research}, pages 1554--1565. {PMLR}.

\bibitem[{Cheng et~al.(2021)Cheng, Chen, Liu, Chang, Hsieh, and Das}]{sprout}
Minhao Cheng, Pin{-}Yu Chen, Sijia Liu, Shiyu Chang, Cho{-}Jui Hsieh, and Payel
  Das. 2021.
\newblock \href {https://ojs.aaai.org/index.php/AAAI/article/view/16874}
  {Self-progressing robust training}.
\newblock In \emph{Thirty-Fifth {AAAI} Conference on Artificial Intelligence,
  {AAAI} 2021, Thirty-Third Conference on Innovative Applications of Artificial
  Intelligence, {IAAI} 2021, The Eleventh Symposium on Educational Advances in
  Artificial Intelligence, {EAAI} 2021, Virtual Event, February 2-9, 2021},
  pages 7107--7115. {AAAI} Press.

\bibitem[{Clark et~al.(2019)Clark, Lee, Chang, Kwiatkowski, Collins, and
  Toutanova}]{clark2019boolq}
Christopher Clark, Kenton Lee, Ming-Wei Chang, Tom Kwiatkowski, Michael
  Collins, and Kristina Toutanova. 2019.
\newblock Boolq: Exploring the surprising difficulty of natural yes/no
  questions.
\newblock \emph{arXiv preprint arXiv:1905.10044}.

\bibitem[{Cohen et~al.(2019)Cohen, Rosenfeld, and Kolter}]{cohen2019certified}
Jeremy Cohen, Elan Rosenfeld, and Zico Kolter. 2019.
\newblock \href {https://proceedings.mlr.press/v97/cohen19c.html} {Certified
  adversarial robustness via randomized smoothing}.
\newblock In \emph{Proceedings of the 36th International Conference on Machine
  Learning}, volume~97 of \emph{Proceedings of Machine Learning Research},
  pages 1310--1320. PMLR.

\bibitem[{Dagan et~al.(2005)Dagan, Glickman, and Magnini}]{dagan2005RTE}
Ido Dagan, Oren Glickman, and Bernardo Magnini. 2005.
\newblock The pascal recognising textual entailment challenge.
\newblock In \emph{Machine learning challenges workshop}, pages 177--190.
  Springer.

\bibitem[{Davison et~al.(2019)Davison, Feldman, and Rush}]{DavisonFR19}
Joe Davison, Joshua Feldman, and Alexander~M. Rush. 2019.
\newblock \href {https://doi.org/10.18653/v1/D19-1109} {Commonsense knowledge
  mining from pretrained models}.
\newblock In \emph{Proceedings of the 2019 Conference on Empirical Methods in
  Natural Language Processing and the 9th International Joint Conference on
  Natural Language Processing, {EMNLP-IJCNLP} 2019, Hong Kong, China, November
  3-7, 2019}, pages 1173--1178. Association for Computational Linguistics.

\bibitem[{Devlin et~al.(2019)Devlin, Chang, Lee, and
  Toutanova}]{devlin2018bert}
Jacob Devlin, Ming-Wei Chang, Kenton Lee, and Kristina Toutanova. 2019.
\newblock \href {https://doi.org/10.18653/v1/N19-1423} {{BERT}: Pre-training of
  deep bidirectional transformers for language understanding}.
\newblock In \emph{Proceedings of the 2019 Conference of the North {A}merican
  Chapter of the Association for Computational Linguistics: Human Language
  Technologies, Volume 1 (Long and Short Papers)}, pages 4171--4186,
  Minneapolis, Minnesota. Association for Computational Linguistics.

\bibitem[{Ebrahimi et~al.(2018)Ebrahimi, Rao, Lowd, and
  Dou}]{ebrahimi2017hotflip}
Javid Ebrahimi, Anyi Rao, Daniel Lowd, and Dejing Dou. 2018.
\newblock \href {https://doi.org/10.18653/v1/P18-2006} {{H}ot{F}lip: White-box
  adversarial examples for text classification}.
\newblock In \emph{Proceedings of the 56th Annual Meeting of the Association
  for Computational Linguistics (Volume 2: Short Papers)}, pages 31--36,
  Melbourne, Australia. Association for Computational Linguistics.

\bibitem[{Gan et~al.(2020)Gan, Chen, Li, Zhu, Cheng, and Liu}]{gan2020vl}
Zhe Gan, Yen-Chun Chen, Linjie Li, Chen Zhu, Yu~Cheng, and Jingjing Liu. 2020.
\newblock \href
  {https://proceedings.neurips.cc/paper/2020/file/49562478de4c54fafd4ec46fdb297de5-Paper.pdf}
  {Large-scale adversarial training for vision-and-language representation
  learning}.
\newblock In \emph{Advances in Neural Information Processing Systems},
  volume~33, pages 6616--6628. Curran Associates, Inc.

\bibitem[{Goodfellow et~al.(2015)Goodfellow, Shlens, and
  Szegedy}]{goodfellow2014explaining}
Ian~J. Goodfellow, Jonathon Shlens, and Christian Szegedy. 2015.
\newblock \href {http://arxiv.org/abs/1412.6572} {Explaining and harnessing
  adversarial examples}.
\newblock In \emph{3rd International Conference on Learning Representations,
  {ICLR} 2015}.

\bibitem[{Gordon et~al.(2012)Gordon, Kozareva, and Roemmele}]{gordon2012copa}
Andrew Gordon, Zornitsa Kozareva, and Melissa Roemmele. 2012.
\newblock Semeval-2012 task 7: Choice of plausible alternatives: An evaluation
  of commonsense causal reasoning.
\newblock In \emph{* SEM 2012: The First Joint Conference on Lexical and
  Computational Semantics--Volume 1: Proceedings of the main conference and the
  shared task, and Volume 2: Proceedings of the Sixth International Workshop on
  Semantic Evaluation (SemEval 2012)}, pages 394--398.

\bibitem[{Haviv et~al.(2021)Haviv, Berant, and Globerson}]{HavivBG21}
Adi Haviv, Jonathan Berant, and Amir Globerson. 2021.
\newblock \href {https://doi.org/10.18653/v1/2021.eacl-main.316} {Bertese:
  Learning to speak to {BERT}}.
\newblock In \emph{Proceedings of the 16th Conference of the European Chapter
  of the Association for Computational Linguistics: Main Volume, {EACL} 2021},
  pages 3618--3623. Association for Computational Linguistics.

\bibitem[{He et~al.(2021)He, Liu, Gao, and Chen}]{he2020deberta}
Pengcheng He, Xiaodong Liu, Jianfeng Gao, and Weizhu Chen. 2021.
\newblock \href {https://openreview.net/forum?id=XPZIaotutsD} {Deberta:
  decoding-enhanced bert with disentangled attention}.
\newblock In \emph{9th International Conference on Learning Representations,
  {ICLR} 2021, Virtual Event, Austria, May 3-7, 2021}. OpenReview.net.

\bibitem[{Jiang et~al.(2020)Jiang, Xu, Araki, and Neubig}]{jiang2020can}
Zhengbao Jiang, Frank~F. Xu, Jun Araki, and Graham Neubig. 2020.
\newblock \href {https://doi.org/10.1162/tacl\_a\_00324} {How can we know what
  language models know}.
\newblock \emph{Trans. Assoc. Comput. Linguistics}, 8:423--438.

\bibitem[{Jin et~al.(2019)Jin, Jin, Zhou, and Szolovits}]{textfooler}
Di~Jin, Zhijing Jin, Joey~Tianyi Zhou, and Peter Szolovits. 2019.
\newblock \href {http://arxiv.org/abs/1907.11932} {Is {BERT} really robust?
  natural language attack on text classification and entailment}.
\newblock \emph{CoRR}, abs/1907.11932.

\bibitem[{Khashabi et~al.(2018)Khashabi, Chaturvedi, Roth, Upadhyay, and
  Roth}]{multirc}
Daniel Khashabi, Snigdha Chaturvedi, Michael Roth, Shyam Upadhyay, and Dan
  Roth. 2018.
\newblock Looking beyond the surface: A challenge set for reading comprehension
  over multiple sentences.
\newblock In \emph{Proceedings of the 2018 Conference of the North American
  Chapter of the Association for Computational Linguistics: Human Language
  Technologies, Volume 1 (Long Papers)}, pages 252--262.

\bibitem[{Kong et~al.(2022)Kong, Li, Ding, Wu, Zhu, Ghanem, Taylor, and
  Goldstein}]{kong2022robust}
Kezhi Kong, Guohao Li, Mucong Ding, Zuxuan Wu, Chen Zhu, Bernard Ghanem, Gavin
  Taylor, and Tom Goldstein. 2022.
\newblock \href {https://doi.org/10.1109/CVPR52688.2022.00016} {Robust
  optimization as data augmentation for large-scale graphs}.
\newblock In \emph{{IEEE/CVF} Conference on Computer Vision and Pattern
  Recognition, {CVPR} 2022, New Orleans, LA, USA, June 18-24, 2022}, pages
  60--69. {IEEE}.

\bibitem[{Lan et~al.(2020)Lan, Chen, Goodman, Gimpel, Sharma, and
  Soricut}]{lan2019albert}
Zhenzhong Lan, Mingda Chen, Sebastian Goodman, Kevin Gimpel, Piyush Sharma, and
  Radu Soricut. 2020.
\newblock \href {https://openreview.net/forum?id=H1eA7AEtvS} {{ALBERT:} {A}
  lite {BERT} for self-supervised learning of language representations}.
\newblock In \emph{8th International Conference on Learning Representations,
  {ICLR} 2020, Addis Ababa, Ethiopia, April 26-30, 2020}. OpenReview.net.

\bibitem[{Lester et~al.(2021)Lester, Al-Rfou, and Constant}]{lester2021power}
Brian Lester, Rami Al-Rfou, and Noah Constant. 2021.
\newblock \href {https://doi.org/10.18653/v1/2021.emnlp-main.243} {The power of
  scale for parameter-efficient prompt tuning}.
\newblock In \emph{Proceedings of the 2021 Conference on Empirical Methods in
  Natural Language Processing}, pages 3045--3059, Online and Punta Cana,
  Dominican Republic. Association for Computational Linguistics.

\bibitem[{Levesque et~al.(2012)Levesque, Davis, and
  Morgenstern}]{levesque2012WSC}
Hector Levesque, Ernest Davis, and Leora Morgenstern. 2012.
\newblock The winograd schema challenge.
\newblock In \emph{Thirteenth international conference on the principles of
  knowledge representation and reasoning}.

\bibitem[{Li et~al.(2020)Li, Ma, Guo, Xue, and Qiu}]{li-etal-2020-bert-attack}
Linyang Li, Ruotian Ma, Qipeng Guo, Xiangyang Xue, and Xipeng Qiu. 2020.
\newblock \href {https://doi.org/10.18653/v1/2020.emnlp-main.500}
  {{BERT}-{ATTACK}: Adversarial attack against {BERT} using {BERT}}.
\newblock In \emph{Proceedings of the 2020 Conference on Empirical Methods in
  Natural Language Processing (EMNLP)}, pages 6193--6202, Online. Association
  for Computational Linguistics.

\bibitem[{Li and Liang(2021)}]{li2021prefix}
Xiang~Lisa Li and Percy Liang. 2021.
\newblock \href {https://doi.org/10.18653/v1/2021.acl-long.353} {Prefix-tuning:
  Optimizing continuous prompts for generation}.
\newblock In \emph{Proceedings of the 59th Annual Meeting of the Association
  for Computational Linguistics and the 11th International Joint Conference on
  Natural Language Processing, {ACL/IJCNLP} 2021, (Volume 1: Long Papers),
  Virtual Event, August 1-6, 2021}, pages 4582--4597. Association for
  Computational Linguistics.

\bibitem[{Liu et~al.(2021{\natexlab{a}})Liu, Yuan, Fu, Jiang, Hayashi, and
  Neubig}]{liu2021pre}
Pengfei Liu, Weizhe Yuan, Jinlan Fu, Zhengbao Jiang, Hiroaki Hayashi, and
  Graham Neubig. 2021{\natexlab{a}}.
\newblock \href {http://arxiv.org/abs/2107.13586} {Pre-train, prompt, and
  predict: {A} systematic survey of prompting methods in natural language
  processing}.
\newblock \emph{CoRR}, abs/2107.13586.

\bibitem[{Liu et~al.(2022)Liu, Ji, Fu, Tam, Du, Yang, and Tang}]{ptuningv2}
Xiao Liu, Kaixuan Ji, Yicheng Fu, Weng Tam, Zhengxiao Du, Zhilin Yang, and Jie
  Tang. 2022.
\newblock \href {https://doi.org/10.18653/v1/2022.acl-short.8} {{P}-tuning:
  Prompt tuning can be comparable to fine-tuning across scales and tasks}.
\newblock In \emph{Proceedings of the 60th Annual Meeting of the Association
  for Computational Linguistics (Volume 2: Short Papers)}, pages 61--68,
  Dublin, Ireland. Association for Computational Linguistics.

\bibitem[{Liu et~al.(2021{\natexlab{b}})Liu, Zheng, Du, Ding, Qian, Yang, and
  Tang}]{ptuning2021}
Xiao Liu, Yanan Zheng, Zhengxiao Du, Ming Ding, Yujie Qian, Zhilin Yang, and
  Jie Tang. 2021{\natexlab{b}}.
\newblock \href {http://arxiv.org/abs/2103.10385} {{GPT} understands, too}.
\newblock \emph{CoRR}, abs/2103.10385.

\bibitem[{Liu et~al.(2019)Liu, Ott, Goyal, Du, Joshi, Chen, Levy, Lewis,
  Zettlemoyer, and Stoyanov}]{liu2019roberta}
Yinhan Liu, Myle Ott, Naman Goyal, Jingfei Du, Mandar Joshi, Danqi Chen, Omer
  Levy, Mike Lewis, Luke Zettlemoyer, and Veselin Stoyanov. 2019.
\newblock \href {http://arxiv.org/abs/1907.11692} {Roberta: {A} robustly
  optimized {BERT} pretraining approach}.
\newblock \emph{CoRR}, abs/1907.11692.

\bibitem[{Madry et~al.(2018)Madry, Makelov, Schmidt, Tsipras, and
  Vladu}]{madry2018PGD}
Aleksander Madry, Aleksandar Makelov, Ludwig Schmidt, Dimitris Tsipras, and
  Adrian Vladu. 2018.
\newblock \href {https://openreview.net/forum?id=rJzIBfZAb} {Towards deep
  learning models resistant to adversarial attacks}.
\newblock In \emph{6th International Conference on Learning Representations,
  {ICLR} 2018}. OpenReview.net.

\bibitem[{Morris et~al.(2020)Morris, Yoo, and Qi}]{morris2020textattack}
John Morris, Jin~Yong Yoo, and Yanjun Qi. 2020.
\newblock \href {https://doi.org/10.18653/v1/2020.nlposs-1.18} {{T}ext{A}ttack:
  Lessons learned in designing python frameworks for {NLP}}.
\newblock In \emph{Proceedings of Second Workshop for NLP Open Source Software
  (NLP-OSS)}, pages 126--131. Association for Computational Linguistics.

\bibitem[{Pilehvar and Camacho-Collados(2018)}]{pilehvar2018wic}
Mohammad~Taher Pilehvar and Jose Camacho-Collados. 2018.
\newblock Wic: the word-in-context dataset for evaluating context-sensitive
  meaning representations.
\newblock \emph{arXiv preprint arXiv:1808.09121}.

\bibitem[{Raffel et~al.(2020)Raffel, Shazeer, Roberts, Lee, Narang, Matena,
  Zhou, Li, and Liu}]{raffel2020exploring}
Colin Raffel, Noam Shazeer, Adam Roberts, Katherine Lee, Sharan Narang, Michael
  Matena, Yanqi Zhou, Wei Li, and Peter~J. Liu. 2020.
\newblock \href {http://jmlr.org/papers/v21/20-074.html} {Exploring the limits
  of transfer learning with a unified text-to-text transformer}.
\newblock \emph{Journal of Machine Learning Research(JMLR)}.

\bibitem[{Ren et~al.(2019)Ren, Deng, He, and Che}]{ren2019generating}
Shuhuai Ren, Yihe Deng, Kun He, and Wanxiang Che. 2019.
\newblock \href {https://doi.org/10.18653/v1/P19-1103} {Generating natural
  language adversarial examples through probability weighted word saliency}.
\newblock In \emph{Proceedings of the 57th Annual Meeting of the Association
  for Computational Linguistics}, pages 1085--1097, Florence, Italy.
  Association for Computational Linguistics.

\bibitem[{Sanh et~al.(2019)Sanh, Debut, Chaumond, and
  Wolf}]{sanh2019distilbert}
Victor Sanh, Lysandre Debut, Julien Chaumond, and Thomas Wolf. 2019.
\newblock \href {http://arxiv.org/abs/1910.01108} {Distilbert, a distilled
  version of {BERT:} smaller, faster, cheaper and lighter}.
\newblock \emph{CoRR}, abs/1910.01108.

\bibitem[{Schick and Sch{\"{u}}tze(2021{\natexlab{a}})}]{schick2020exploiting}
Timo Schick and Hinrich Sch{\"{u}}tze. 2021{\natexlab{a}}.
\newblock \href {https://doi.org/10.18653/v1/2021.eacl-main.20} {Exploiting
  cloze-questions for few-shot text classification and natural language
  inference}.
\newblock In \emph{Proceedings of the 16th Conference of the European Chapter
  of the Association for Computational Linguistics: Main Volume, {EACL} 2021},
  pages 255--269. Association for Computational Linguistics.

\bibitem[{Schick and Sch{\"{u}}tze(2021{\natexlab{b}})}]{fewglue}
Timo Schick and Hinrich Sch{\"{u}}tze. 2021{\natexlab{b}}.
\newblock \href {https://doi.org/10.18653/v1/2021.naacl-main.185} {It's not
  just size that matters: Small language models are also few-shot learners}.
\newblock In \emph{Proceedings of the 2021 Conference of the North American
  Chapter of the Association for Computational Linguistics: Human Language
  Technologies, {NAACL-HLT} 2021, Online, June 6-11, 2021}, pages 2339--2352.
  Association for Computational Linguistics.

\bibitem[{Shoeybi et~al.(2019)Shoeybi, Patwary, Puri, LeGresley, Casper, and
  Catanzaro}]{megaron-LM}
Mohammad Shoeybi, Mostofa Patwary, Raul Puri, Patrick LeGresley, Jared Casper,
  and Bryan Catanzaro. 2019.
\newblock \href {http://arxiv.org/abs/1909.08053} {Megatron-lm: Training
  multi-billion parameter language models using model parallelism}.
\newblock \emph{CoRR}, abs/1909.08053.

\bibitem[{Tram{\`{e}}r et~al.(2018)Tram{\`{e}}r, Kurakin, Papernot, Goodfellow,
  Boneh, and McDaniel}]{tramer2017ensemble}
Florian Tram{\`{e}}r, Alexey Kurakin, Nicolas Papernot, Ian~J. Goodfellow, Dan
  Boneh, and Patrick~D. McDaniel. 2018.
\newblock \href {https://openreview.net/forum?id=rkZvSe-RZ} {Ensemble
  adversarial training: Attacks and defenses}.
\newblock In \emph{6th International Conference on Learning Representations,
  {ICLR} 2018, Vancouver, BC, Canada, April 30 - May 3, 2018, Conference Track
  Proceedings}. OpenReview.net.

\bibitem[{Vaswani et~al.(2017)Vaswani, Shazeer, Parmar, Uszkoreit, Jones,
  Gomez, Kaiser, and Polosukhin}]{BERT}
Ashish Vaswani, Noam Shazeer, Niki Parmar, Jakob Uszkoreit, Llion Jones,
  Aidan~N. Gomez, Lukasz Kaiser, and Illia Polosukhin. 2017.
\newblock \href {http://arxiv.org/abs/1706.03762} {Attention is all you need}.
\newblock \emph{CoRR}, abs/1706.03762.

\bibitem[{Wang et~al.(2019)Wang, Pruksachatkun, Nangia, Singh, Michael, Hill,
  Levy, and Bowman}]{superglue}
Alex Wang, Yada Pruksachatkun, Nikita Nangia, Amanpreet Singh, Julian Michael,
  Felix Hill, Omer Levy, and Samuel Bowman. 2019.
\newblock \href
  {https://proceedings.neurips.cc/paper/2019/file/4496bf24afe7fab6f046bf4923da8de6-Paper.pdf}
  {Superglue: A stickier benchmark for general-purpose language understanding
  systems}.
\newblock In \emph{NeurIPS 2019}.

\bibitem[{Xie et~al.(2020)Xie, Tan, Gong, Wang, Yuille, and
  Le}]{xie2020adversarial}
Cihang Xie, Mingxing Tan, Boqing Gong, Jiang Wang, Alan~L. Yuille, and Quoc~V.
  Le. 2020.
\newblock \href {https://doi.org/10.1109/CVPR42600.2020.00090} {Adversarial
  examples improve image recognition}.
\newblock In \emph{2020 {IEEE/CVF} Conference on Computer Vision and Pattern
  Recognition, {CVPR} 2020, Seattle, WA, USA, June 13-19, 2020}, pages
  816--825. Computer Vision Foundation / {IEEE}.

\bibitem[{Yoo et~al.(2020)Yoo, Morris, Lifland, and Qi}]{yoo2020searching}
Jin~Yong Yoo, John Morris, Eli Lifland, and Yanjun Qi. 2020.
\newblock \href {https://doi.org/10.18653/v1/2020.blackboxnlp-1.30} {Searching
  for a search method: Benchmarking search algorithms for generating {NLP}
  adversarial examples}.
\newblock In \emph{Proceedings of the Third BlackboxNLP Workshop on Analyzing
  and Interpreting Neural Networks for NLP}, pages 323--332, Online.
  Association for Computational Linguistics.

\bibitem[{Yoo and Qi(2021)}]{a2t}
Jin~Yong Yoo and Yanjun Qi. 2021.
\newblock \href {https://doi.org/10.18653/v1/2021.findings-emnlp.81} {Towards
  improving adversarial training of {NLP} models}.
\newblock In \emph{Findings of the Association for Computational Linguistics:
  {EMNLP} 2021, Virtual Event / Punta Cana, Dominican Republic, 16-20 November,
  2021}, pages 945--956. Association for Computational Linguistics.

\bibitem[{Zhu et~al.(2020)Zhu, Cheng, Gan, Sun, Goldstein, and Liu}]{freelb}
Chen Zhu, Yu~Cheng, Zhe Gan, Siqi Sun, Tom Goldstein, and Jingjing Liu. 2020.
\newblock \href {https://openreview.net/forum?id=BygzbyHFvB} {Freelb: Enhanced
  adversarial training for natural language understanding}.
\newblock In \emph{8th International Conference on Learning Representations,
  {ICLR} 2020, Addis Ababa, Ethiopia, April 26-30, 2020}.

\end{thebibliography}
\bibliographystyle{acl_natbib}

\appendix

\newpage

\begin{center}
   \textbf{APPENDIX}
\end{center}

\section{Details of learning rate and prompt length}
\label{sec: details of lr and pl}

\begin{figure*}[!t]
    \centering
    \includegraphics[width=0.99\textwidth]{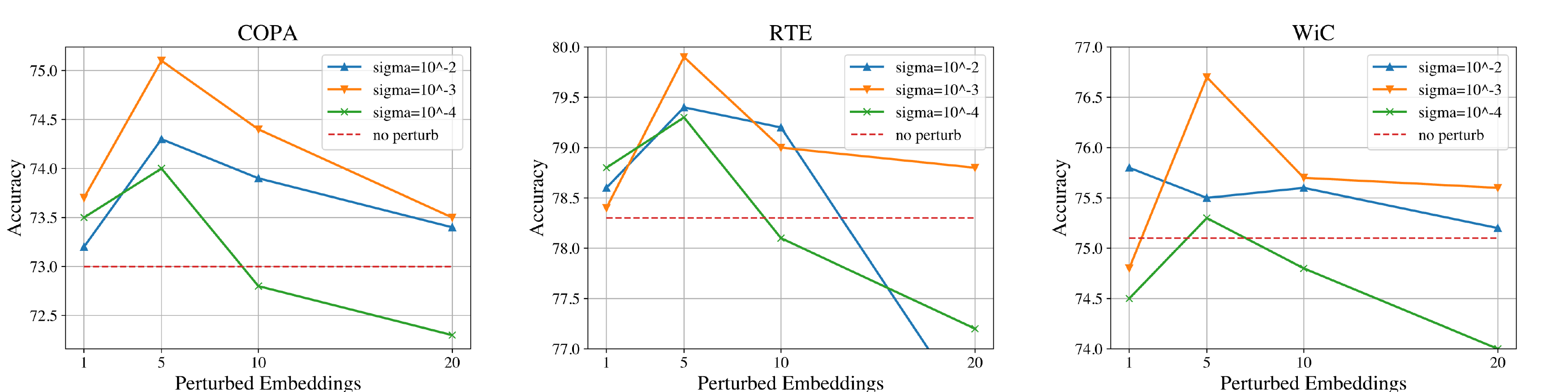}
    \caption{Performance of PTP+RG on SuperGLUE~(WiC, RTE, COPA datasets) with $\sigma$ from \{$10^{-4}$, $10^{-3}$, $10^{-2}$\} and perturbed embeddings from \{1, 5, 10, 20\}. The dashed red line represents the performance of baseline method PT2 with BERT-large as backbone LM. }
    \label{fig:few-rn.}
\end{figure*}

\begin{figure*}[!t]
    \centering
    \includegraphics[width=0.99\textwidth]{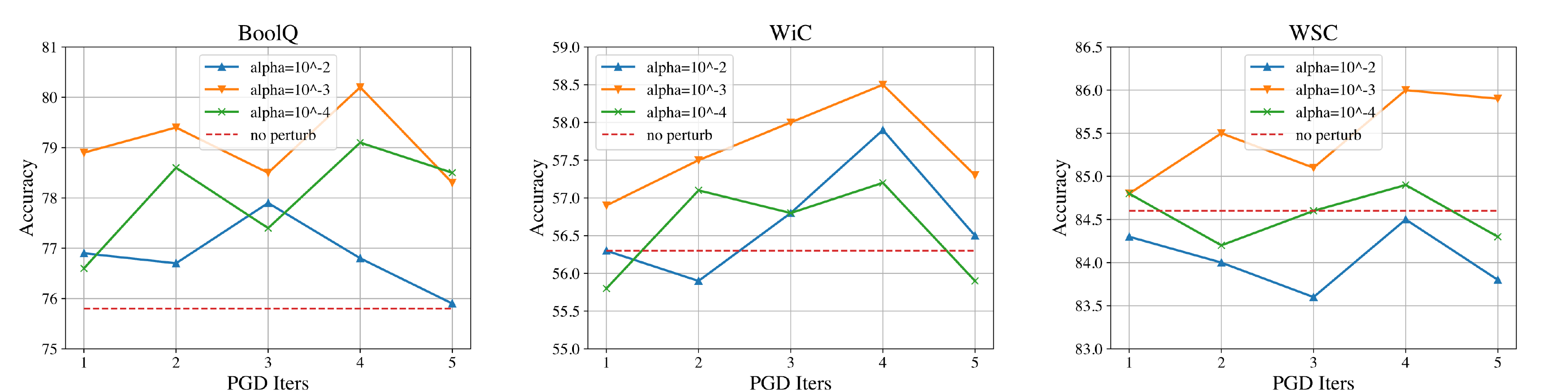}
    \caption{Performance of PTP+PGD on FewGLUE~(WiC, BoolQ, WSC datasets) with different $\alpha$ and PGD iterations. The dashed red line represents the performance of the baseline method PT~\cite{ptuning2021}. It shows the best~$\alpha$ and PGD iterations are $10^{-3}$ and 4, respectively.}
    \label{fig:few-pgd.}
\end{figure*}

\begin{table}[ht]
\centering
\begin{tabular}{ccccc}
\toprule[1pt]
Tasks  & LR1  & LR2  & PL1 & PL2 \\ \hline
BoolQ  & 1e-3 & 5e-3 & 40  & 16  \\
COPA   & 5e-3 & 7e-3 & 16  & 16  \\
RTE    & 1e-2 & 5e-3 & 20  & 128 \\
WSC    & 3e-3 & 7e-3 & 16  & 8   \\
CB     & 7e-3 & 9e-3 & 16  & 16  \\
MultiRC & 1e-4 & 3e-3 & 40  & 20  \\
ReCoRD & 3e-4 & 5e-3 & 16 & 40 \\
WiC    & 1e-4 & 5e-3 & 20  & 16  \\ \bottomrule[1pt]
\end{tabular}
\caption{Prompt Length and Learning Rate details for 8 tasks on SuperGLUE. LR1, PL1: learning rate and prompt length for continuous prompts with BERT-large backbone. LR2 and PL2: learning rate and prompt length for prompts with RoBERTa-large backbone.}
\label{tab: lr and pl for superglue}
\end{table}

\begin{table}[ht]
\centering
\begin{tabular}{cc}
\toprule[1pt]
Tasks   & Learning Rate \\ \hline
BoolQ   & 5e-5          \\
RTE     & 5e-5          \\
WiC     & 1e-5          \\
WSC     & 5e-5          \\
COPA    & 1e-5          \\
MultiRC & 1e-4          \\
CB      & 1e-5          \\ \bottomrule[1pt]
\end{tabular}
\caption{Prompts' Learning Rate details for 7 tasks in FewGLUE.}
\label{fewglue: learning rate}
\end{table}

\par Under fully-supervised settings, the prompt length and learning rate details are presented as Table~\ref{tab: lr and pl for superglue}. Under few-shot learning settings, we report it as Table~\ref{fewglue: learning rate}. The prompt length is exactly the same as the PT~\cite{ptuning2021}, thus we ignore it here.

\begin{table*}[!t]
\centering
\begin{tabular}{c|cccc}
\toprule[1pt]
Tasks   & RTE    & WSC   & WiC & BoolQ \\ \hline
PT   &  1.89    & 1.68   & 1.77 &  1.45       \\
+RG  & 0.81   & 0.78   & 0.91   &  0.56         \\
+A2T & 0.45  & 0.43  & \textbf{0.39}\down{$\downarrow$1.38}  & 0.59             \\
+RM  & 0.68 & 0.95 & 0.87 & 0.65 \\
+PGD & \textbf{0.35}\down{$\downarrow$1.54}   & \textbf{0.31}\down{$\downarrow$1.37}   & 0.47 &  \textbf{0.43}\down{$\downarrow$1.02}  \\ 
\bottomrule[1pt]
\end{tabular}
\caption{The variance of the scores on the dev sets of RTE, COPA and BoolQ from the FewGLUE benchmark. We compute it on 5 runs with different random seeds~(other hyper-parameter are the same). We employ bold font to denote the smallest deviation in each task and blue font to denote the decrease when compared to PT.}
\label{tab:robustness}
\end{table*}

\section{Supplement for Experiment}
\par This section includes the tables and figures as a supplement to our experiment and ablations. Table~\ref{tab:robustness} demonstrates the comparison of the variance of different training methods on FewGLUE benchmark. Table~\ref{tab: few-MultiRC} presents the ablation of our algorithm with RG perturbation on MultiRC~\cite{multirc} task. We show the ablation of PGD perturbation in Table~\ref{tab:full-pgd}. The ablation of RM perturbation is presented as Table~\ref{tab:ablationRM}. Table~\ref{tab:A2T ablation} shows the ablation of our proposed PTP+A2T training algorithm.

\par
\begin{figure*}[!t]
    \centering
    \includegraphics[width=0.8\textwidth]{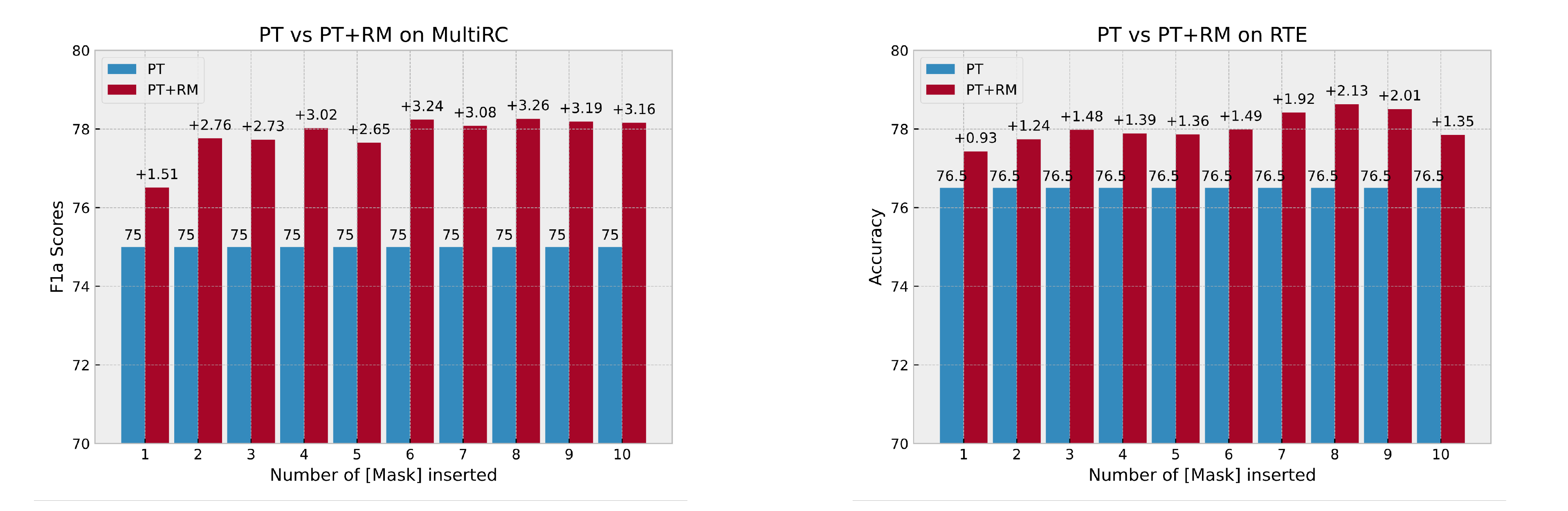}
    \caption{The results of different $i$ in Eq.~(\ref{eq:RM})~(number of [MASK] randomly inserted into input sequence as perturbation). We select MultiRC and RTE tasks with Few-shot setting to report. }
    \label{fig:few-shot setting PT+RM.}
\end{figure*}

 Figure~\ref{fig:few-rn.} shows the ablation of RG perturbations on WiC~\cite{pilehvar2018wic}, RTE~\cite{dagan2005RTE}, and COPA~\cite{gordon2012copa} datasets under fully-supervised learning setting while Figure~\ref{fig:few-pgd.} presents the ablation of our PTP+PGD training method on WiC~\cite{pilehvar2018wic}, BoolQ~\cite{clark2019boolq} and WSC~\cite{levesque2012WSC} datasets under few-shot learning settings. We show the ablation of our proposed PTP+RM algorithm as Figure~\ref{fig:few-shot setting PT+RM.} on FewGLUE benchmark. 

\par All experiments are conducted on servers with RTX A6000 GPUs, each having 48GB of memory.

\begin{table*}[!t]
\centering
\begin{tabular}{c|cccc}
\toprule[1pt]
MultiRC & E=1    & E=5    & E=10   & E=20   \\ \hline
$\sigma$=1e-2        & 75.2 & 74.3\down{$\downarrow$0.7} & 75.8 & 74.7  \\
$\sigma$=1e-3        & 77.3 & \textbf{78.1}\hld{$\uparrow$3.1} & 77.1 & 76.6 \\
$\sigma$=1e-4        & 76.9 & 77.2 & 76.8 & 75.5 \\
\bottomrule[1pt]
\end{tabular}
\caption{Results of different $\sigma$ in RG perturbation~(MultiRC task). The baseline method is PT2 and its performance is~$75.0$. E denotes number of embeddings that are perturbed. We mark the best and the worst.}
\label{tab: few-MultiRC}
\end{table*} 

\begin{table*}[!t]
\centering
\begin{tabular}{c|ccccc}
\toprule[1pt]
 COPA & t=1    & 2    & 3    & 4    & 5    \\ \hline
$\alpha$=1e-2        & 70.3 & 72.1 & 72.0 & 71.0 & 69.0\down{$\downarrow$4.0} \\
1e-3        & 73.4 & 70.8 & 72.9 & 73.5 & 73.2 \\
1e-4        & 73.1 & 73.4 & \textbf{74.7}\hld{$\uparrow$1.7} & 73.8  & 72.5 \\ 
\bottomrule[1pt]
\end{tabular}

\caption{Results of PTP+PGD on fully-supervised COPA dataset. The baseline method is PT2~\cite{ptuningv2}, whose accuracy is~$73.0$~. The backbone LM employed is BERT-large. $\alpha$~is the perturbation size while $t$~is PGD iterations~(see Eq.~(\ref{eq:pt-pgd})). }
\label{tab:full-pgd}
\end{table*}

\begin{table*}[!t]
\centering
\begin{tabular}{c|c|cccccccccc}
\toprule[1pt]
Dataset & LM & 1    & 2    & 3    & 4    & 5    & 6 & 7 & 8 & 9 & 10 \\ \hline
BoolQ   &   BERT   & -0.12 & +0.50 & \textbf{+1.57} & +0.65 & +1.35 &  +1.14 & +1.12 & +0.24  &  +0.25  &  -0.38   \\
(Full)  &   RoBERTa & +0.31 & +0.76 & \textbf{+1.10} & +0.88 & +0.69 &  -0.36  &  +0.19  & -0.43  & -0.67  & -0.36   \\
\bottomrule[1pt]
\end{tabular}
\caption{The results of different $i$ in Eq.~(\ref{eq:RM})~(number of [MASK] randomly inserted into input sequence as perturbation)~. We select BoolQ with fully-supervised settings to report. The reported increase or decrease is compared to the baseline method PT2.}
\label{tab:ablationRM}
\end{table*}

\begin{table*}[!t]
\centering
\begin{tabular}{c|cccc}
\toprule[1pt]
Cosine Sim    & 0.2  & 0.4  & 0.6  & 0.8  \\ \hline
BoolQ(Full)   & -0.83 & -0.46 & +0.61 & +0.52 \\
BoolQ(Few)    &  -0.96 &   -0.13   &  +0.78   & +0.84     \\
MultiRC(Full) &  -0.77 &   -0.35   &   +0.25   & +0.71     \\
MultiRC(Few)  &   -0.59  &  -0.54    &   +1.03   & +0.68    \\
\bottomrule[1pt]
\end{tabular}
\caption{The results of different minimum cosine similarity in A2T perturbation. Full: fully-supervised learning setting. Few: few-shot learning setting.}
\label{tab:A2T ablation}
\end{table*}

% \clearpage
% \section{Limitations}
% \par In this paper, we introduce the PTP algorithm, which enhances the stability and performance of prompt tuning. It should be noted, however, that our approach is exclusively relevant to continuous prompt tuning, and cannot be used to improve other models like GPT-3 or ChatGPT, which are only accessible via their APIs. Furthermore, our algorithm has a limitation in that we do not provide a theoretical analysis.

% \section{Rebuttal}
% \subsection{Reply to Reviewer NDBS}
% Thanks for your encouraging and insightful comments. We will continue to polish the paper. Below, we provide detailed responses to your questions. \\
% Q1: About the comparison of the variance to the vanilla fine-tuning.\\
% A1: Thanks for the great suggestions. We show the results in the following table. It shows the vanilla fine-tuning achieves around 10X smaller variance compared with the vanilla training. We will add this result to Table 1 in the revision. 
% \begin{table}[]
%     \centering
%     \begin{tabular}{cccc}
%     \toprule[1pt]
%       & RTE  & BoolQ &  WiC  \\ \midrule[0.75pt]
%   Var & 0.12 & 0.09 & 0.11 \\ \bottomrule[1pt]
%     \end{tabular}
%     \caption{Training variance: Vanilla fine-tuning.}
%     \label{tab:my_label}
% \end{table}

\end{document}